\documentclass[runningheads]{llncs}
\usepackage[T1]{fontenc}
\usepackage{graphicx}
\usepackage{amsmath, amssymb}
\usepackage{tikz}
\usetikzlibrary{calc}
\usetikzlibrary{positioning}
\usetikzlibrary{shapes.geometric}
\usetikzlibrary{fadings}
\usepackage{graphicx, tikz}
\usepackage{subcaption}
\usepackage{xcolor}
\usepackage{hyperref}
\hypersetup{
    colorlinks=true,       
    linkcolor=blue,
    citecolor=blue,   
    urlcolor=blue 
}
\usepackage{color}

\usepackage{multicol}

\begin{document}
\title{Evaluating the Generalizability and Explainability of Firn Image Classification with Topological Features\thanks{Day, Keegan, and Jester were partially supported by the Army Research Office under Grant Number W911NF-20-1-0131. The views and conclusions contained in this document are those of the authors and should not be interpreted as representing the official policies, either expressed or implied, of the Army Research Office or the U.S. Government. The U.S. Government is authorized to reproduce and distribute reprints for Government purposes not-withstanding any copyright notation herein. Dimino and Weighill were partially supported by NSF REU Site: Complex Data Analysis using Statistical and Machine Learning Tools (DMS-1950549).}}
\titlerunning{Evaluating Firn Image Classification with Topological Features}
%
\author{Jesse Dimino\inst{1}\orcidID{0009-0008-1776-3780} \and
Matt Jester\inst{2}
\and
Kaitlin Keegan\inst{3}\orcidID{0000-0002-1241-2681} \and
Thomas Weighill\inst{4}\orcidID{0000-0003-2614-0979} \and
Sarah Day\inst{5}\orcidID{0009-0009-8047-0187} 
}
\authorrunning{J.~Dimino et al.}
%
\institute{Department of Mathematics, CUNY College of Staten Island, Staten Island, NY 10314, United States
\email{jesse.dimino@csi.cuny.edu}\\
\and
Division of Student Success, University of North Carolina at Greensboro, Greensboro, NC 27412, United States
\email{mwjester@uncg.edu} \\
\and
Nevada Geosciences, Department of Geological Sciences and Engineering, University of Nevada, Reno, Reno, NV 89557, United States
\email{kkeegan@unr.edu} \\
\and 
Department of Mathematics and Statistics, University of North Carolina at Greensboro, Greensboro, NC 27412, United States \email{t\_weighill@uncg.edu} \\
\and
Department of Mathematics, William \& Mary, Williamsburg, VA 23185, United States \email{sldayx@wm.edu}\\
}
\maketitle              
\begin{abstract}
In this paper we evaluate the performance of topological features for generalizable and explainable classification of firn image data, with the broader goal of understanding the advantages, pitfalls, and trade-offs in topological featurization. Firn refers to layers of granular snow within glaciers that haven't yet been compressed into ice. This compaction process imposes distinct topological and geometric structure on firn that varies with depth within the firn column. However, the limited and highly correlated training data available results in naive classifiers (such as convolutional neural networks) overfitting to orientation-dependent features. This motivates us to evaluate topological data analysis (TDA) as a way to design features which capture the shape of the firn data while retaining a large degree of invariance.
We evaluate two classes of topological features, sublevel set features and distance transform features, together with persistence curves, to predict sample depth from firn images. A range of challenging training-test scenarios reveals that no one choice of method dominates in all categories. In particular, we see a tradeoff between accuracy on the one hand, and generalizability and explainability (via feature importance) on the other. 
\end{abstract}

\section{Introduction}

Image classification is a well-studied area of machine learning with important real-world applications. The classification of textures presents a specific challenge, namely that a well-tuned classifier should be roughly invariant under various rigid transformations of the image. For example, a patch of sand should not change its classification label if rotated $180^\circ$, while an image of a 6 can become an image of a 9 under such a transformation. This invariance can be incorporated into the machine learning pipeline in various ways. One could add transformed versions of each image to the training set (data augmentation), or one could force the required invariance in the architecture of the classifier itself (invariant machine learning~\cite{wood1996representation,cahill2024group}). We focus on a third option, namely \emph{invariant featurization} -- transforming the data into vectors in an invariant way before feeding them into a standard machine learning pipeline.

For this purpose we draw on the well-suited family of methods called topological data analysis (TDA). TDA is a cluster of data analysis methods based on theory from algebraic topology. One key tool in TDA is \emph{persistent homology} which tracks the appearance and disappearance of topological features (such as connected components and holes) over a range of scales or thresholds. Since its introduction by Edelsbrunner, Letscher and Zomorodian in 2000, persistent homology has had wide-ranging applications in areas such as geospatial data, time series data, image classification, medical diagnostics, dimension reduction, deep learning, and more. Most relevant to our work are applications to texture classification in \cite{chung2020smooth,chung2022persistence}.

\begin{figure}[h]
    \centering
    \includegraphics[width = \textwidth]{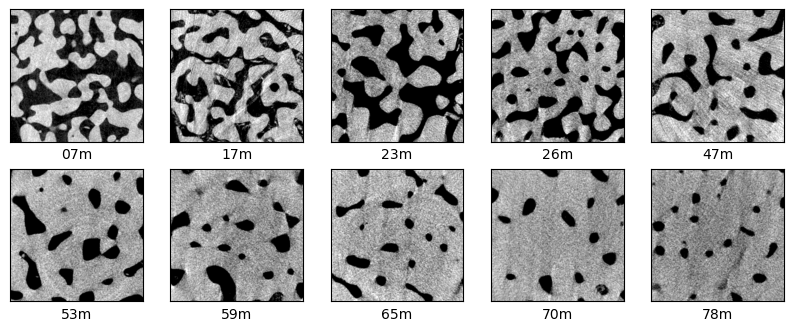}
    \caption{One sample image per depth for the firn dataset. Gray areas represent ice, while the black areas show the pore (air) space.}
    \label{fig:FirnDepths}
\end{figure}

In this paper, we undertake a particular image classification task on geological data to evaluate and compare the performance of various persistent homology featurizations. Specifically, our dataset consists of firn image data obtained from X-ray micro-computed tomography (micro-CT) scans of shallow ice cores. This dataset has a number of unique features, which are visible in the examples in Figure~\ref{fig:FirnDepths}. 
\begin{itemize}
    \item Each image is grayscale and while each is roughly divided into two regions --- pore space (shown in black) and ice space (shown in gray) --- there is enough variation and noise in each to make extracting a binary image non-trivial. 
    \item The pore space and ice space complex (as opposed to the granular noise) can be viewed as a medium-scale texture in that the pattern does not have any well-defined orientation or boundary.  
    \item Even within the sample images, a clear trend can be noticed as depth increases: the pore space goes from a connected complex taking up roughly half of the image to a set of small bubbles.
\end{itemize}

Each sample in our dataset comes with a label indicating the depth (on the ice core) at which the scan was taken, with ten possible values corresponding to ten sampling depths. Our classification task is to predict depth from an image. Firn microstructure evolves systematically with depth through compaction, sintering, and metamorphism, and a long term goal is to apply a depth classifier trained on this ice core to micro-CT images from other cores. Comparing predicted depths with the samples' measured depths would reveal differences in the rate and trajectory of microstructural evolution among sites. It may be, for example, that the site conditions that produced the structure in the $26$m sample in our core corresponds to a greater or shallower depth in a core formed under different temperature and accumulation conditions. Such offsets could therefore provide insight into how site-specific environmental conditions influence firn evolution. Furthermore, an explainable classifier that identifies the image features most influential to its predictions could reveal which aspects of firn microstructure distinguish depths and sites, offering additional physical insight into processes governing firn densification.

Persistent homology features, which are invariant under many different transformations, are naturally suited to this task since they can help mitigate overfitting (say to specific locations within each image) and promote generalizability and robustness. Indeed, TDA methods have already been used to study this dataset~\cite{chung2018topological,hu2021two,belotti2021topological,lawson2022density}. Our study involves two methods of persistent homology featurization: \emph{sublevel set filtration}, which computes topological features at various pixel intensity thresholds, and \emph{distance transforms}, which computes topological features by slowly thickening up the ice space region of the image. 

We set up a range of different prediction and classification tasks, aiming to find a featurization which is accurate, robust to various manipulations of the test set, and which can extrapolate to out-of-sample depth values. Ultimately, we find that none of the featurizations we consider satisfy all these requirements perfectly. Instead, our experiments reveal an interesting web of trade-offs between our desiderata, and insights into the use of topological features on firn images and similar datasets.

We now briefly outline the rest of the paper. We begin in the next section with some necessary background including a more detailed description of the dataset and its geological properties. We discuss the challenge of training a generalizable classifier on this dataset in Section~\ref{sec:cnn}, using a simple CNN as an example. We then outline our classification methods and experiments in Section \ref{sec:methods}, and give some results and discussion in Section \ref{sec:results}. We examine the feature importance outputs of the classifiers in Section~\ref{sec:featureimportance} and discuss their implications. We conclude the paper with a summary of the lessons learned during the process.

\section{Background}

\subsection{Firn Image Data}

Firn is a porous medium that forms in polar regions where snow does not completely melt during the summer season; instead, it accumulates, densifies, and ultimately becomes glacial ice. Firn comprises a matrix of dense snow with an interconnected pore space, which undergoes densification due to the increasing pressure from the accumulation of snowfall on the surface. The zone at the top of glaciers and ice sheets where the firn is densifying is referred to as the firn column. As depth increases, the ice matrix grows and the pore space constricts until the pores close off into individual bubbles within the ice at the bottom of the firn column (Figure \ref{fig:FirnDepths}).

The process of densification and its impact on the underlying microstructure of firn, has important implications on many glaciological applications. Recent studies suggest that microstructural information is needed to improve empirical models of firn densification ~\cite{lundin2017firn}, but the technique generates large amounts of data with unique challenges. Additionally, there are no universal image processing and analysis methods for extracting useful microstructural information from the micro-CT data. TDA is well-suited for improving firn microstructure classification, since topological features are invariant under rotation, translation and other rigid transformations which should be viewed as a result of changes in the sampling technique and not the underlying geological data.

The firn micro-CT data used in this study were collected by Kaitlin Keegan from a core collected at NEEM, Greenland, and they were processed with a Skyscan 1172 micro-CT scanner housed within a cold laboratory. Each scan produced a stack of approximately 900 2-D cross-sectional images that represent the original 3-D sample. The cross-sectional images, or slices, are $500 \times 500$ images grayscale where gray pixels represent the ice-space and black pixels represent the pore-space of the firn. Pixel values range from $0$ (black) to $255$ (white). For each core, every 5\textsuperscript{th} slice was subsampled. We investigate the microstructure of samples from ten sample depths ranging from 7m to 78 m, roughly spanning the entire firn column at that site. Figure \ref{fig:FirnDepths} shows one cross-sectional image from each sample depth. Note how the amount of pore space decreases as the depth increases. The result is a transition from a connected, curved pore space to a collection of bubbles as depth increases. 

\subsection{Persistent Homology}\label{sec:homology}

For an introduction to the full theory of persistent homology we invite the reader to the paper \cite{otter2017roadmap} or to one of the recent book-length treatments such as \cite{carlsson2021topological}. In this section we give a brief overview. Persistent homology begins with a nested sequence of spaces
$$
X_{0}\subseteq X_{1} \subseteq \cdots 
$$
In our case, each $X_i$ will represent a set of pixels in a fixed image, produced in one of two ways (see Section \ref{sec:methods}), which grows as $i$ increases. As the set of pixels grows, topological features such as connected components ($0$-th dimensional features) and holes ($1$-dimensional features) appear and disappear, and they can be tracked in a well-defined way using the cubical complex implementation of persistent homology in~\cite{wagner2011efficient}. Each feature has a \emph{birth} and \emph{death} time. For each dimension, every birth-death pair is represented as a point $(b,d)$ in the plane to create a \emph{persistence diagram}. Note that we always have $b < d$, and it is customary to include the $y=x$ line in persistence diagrams. 




\subsection{Persistence Curves}\label{sec:curves}

In an ideal world, raw persistence diagrams could serve as the input to a classifier. However, most well-established and efficient supervised machine learning algorithms require vectors as input. Converting persistence diagrams into vectors is unsurprisingly an area of active research. Some standard choices are persistence landscapes \cite{bubenik2015statistical} and persistence images~\cite{adams2017persistence}. For our study, we utilize persistence curves, based on their usefulness for texture classification in \cite{chung2022persistence}. Persistence curves are a general family of methods for producing vectors from persistence diagrams. We select two for this paper, namely the Betti curve and the Gaussian persistence curve. Other choices are possible (for example, lifespan curves), but we choose two curves which represent an interesting trade-off between interpretability and stability. Indeed, the Betti curve has a simple interpretation in terms of counting topological features (as we will see below), but is not as smooth the Gaussian persistence curve~\cite{chung2026gaussian}. 

\begin{definition}[Betti curve]
  Consider a $k\textsuperscript{th}$-dimensional persistence diagram $P = \{(b_0,d_0), (b_1,d_1), \ldots, (b_n,d_n) \}$. For $t \in \mathbb{R}$, consider the subset of the plane given by
$$
F_t = \{(x,y) \in \mathbb{R}^2 \mid x \leq t < y\}
$$
which we call a \emph{fundamental box} at threshold $t$. We define the \emph{$k\textsuperscript{th}$ Betti curve} to be the function
$$
\beta_k(t) = \#(F_t \cap P)
$$
where $\#$ counts the number of elements.   
\end{definition}

\begin{figure}[h]
    \centering
        \begin{subfigure}{0.48\linewidth} 
        \includegraphics[width= \linewidth]{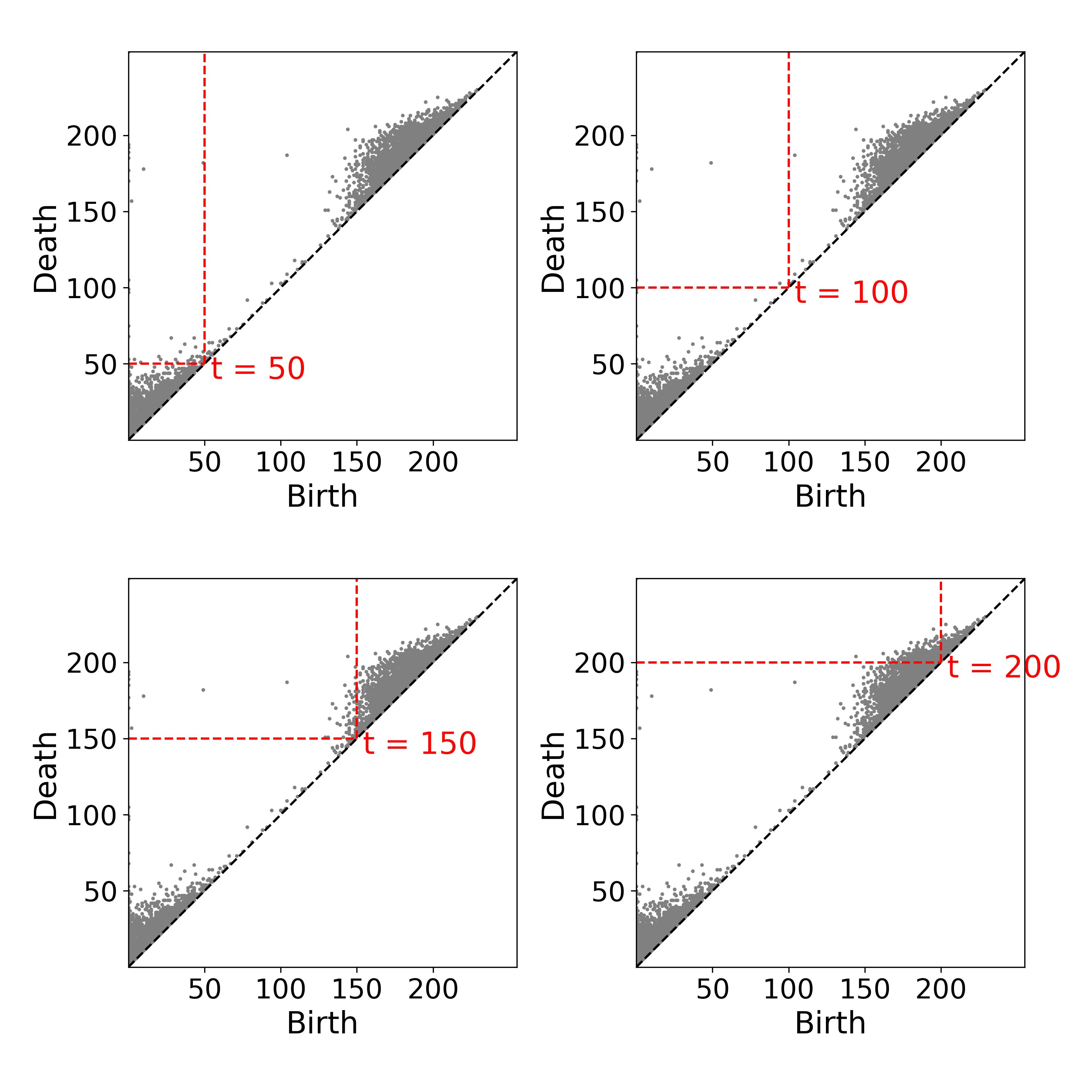}
        \caption{Fundamental box}
        \label{fig:SLOCirclePC}
    \end{subfigure}
    \hfill 
    \begin{subfigure}{0.48\linewidth} 
        \includegraphics[width=\linewidth]{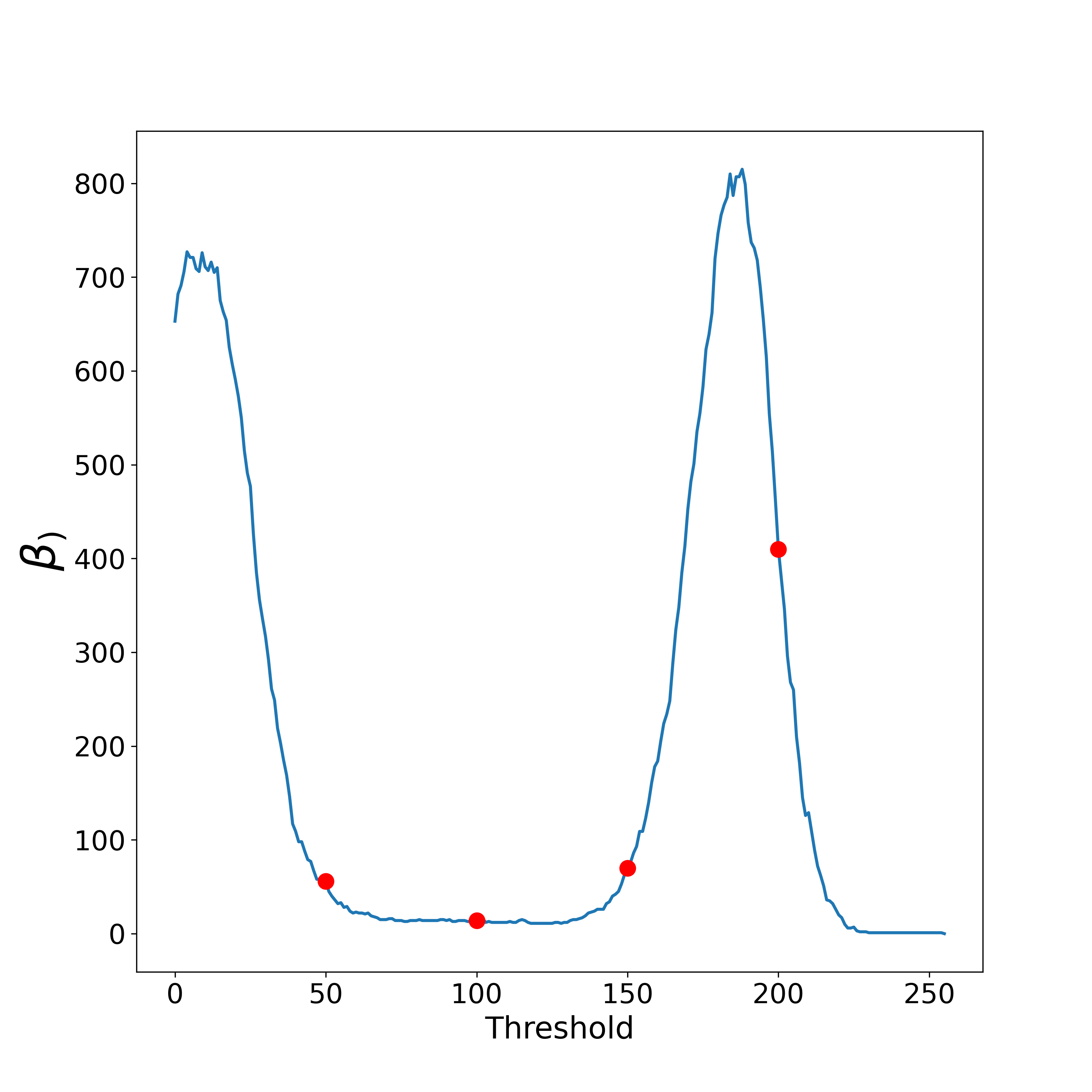}
        \caption{Betti Curve}
        \label{fig:KMCirclesPC}
    \end{subfigure}
    \caption{Computing the $0\textsuperscript{th}$ Betti curve (b) from the $0\textsuperscript{th}$-dimensional persistence diagram (a). The red dashed lines indicate the extent of the fundamental box at each threshold $t$, with these thresholds marked with a red dot on the Betti curve.}
    \label{fig:box}
\end{figure}

See Figure~\ref{fig:box} for a visual exposition of how the Betti curve is created from a persistence diagram. If $t$ ranges over a finite set $\{0,1,\ldots,N\}$, then we can view $\beta_k$ as a vector in $\mathbb{R}^{N}$. In this paper, we always concatenate the $\beta_0$ and $\beta_1$ curves to obtain a vector $\mathbf{v}_{\textrm{Betti}} \in \mathbb{R}^{2N}$. By the Fundamental Lemma of Persistent Homology~\cite[p.~118]{edelsbrunner2022computational}, $\beta_k(t)$ is the $k\textsuperscript{th}$ Betti number of the sublevel set given by the threshold $t$, hence the name. Recall that the $0\textsuperscript{th}$ Betti number counts connected components while the $1\textsuperscript{st}$ Betti number counts holes. In summary: the Betti curves $\beta_0$ and $\beta_1$ encode the changing topology of a sequence in a straightforward way: $\beta_0(t)$ and $\beta_1(t)$ are the number of connected components and holes respectively in $X_t$. 

\begin{definition}[Gaussian persistence curve]
  Consider a $k\textsuperscript{th}$-dimensional persistence diagram $P = \{(b_0,d_0), (b_1,d_1), \ldots, (b_n,d_n) \}$. Given a weighting function $\kappa: \mathbb{R}^2 \to [0,\infty)$ satisfying $\kappa(x,x) = 0$ for all $x \in \mathbb{R}$ and a $2 \times 2$ covariance matrix $\mathbf{\Sigma}$, we define the $k\textsuperscript{th}$-dimensional persistence surface to be the function
  $$
  \rho(P) = \sum_{(b,d) \in P} \kappa(b,d) g_{(b,d), \mathbf{\Sigma}}
  $$
  where $g_{(b,d), \mathbf{\Sigma}}$ is a bivariate normal distribution with mean $(b,d)$ and covariance matrix $\mathbf{\Sigma}$. We define the \emph{$k\textsuperscript{th}$ Gaussian persistence curve} to be the function
$$
\gamma_k(t) = \int_{F_t} \rho(P) dxdy
$$
where $F_t$ is the fundamental box at threshold $t$.   
\end{definition}

The Gaussian persistence curve can be viewed as a smoothed and re-weighted version of the Betti curve: points are replaced by Gaussians, an idea originating with persistence images~\cite{adams2017persistence}, and points far from the diagonal are up-weighted using $\kappa$. In our experiments, we use $\kappa(b,d) = \frac{d-b}{\sum_{(b',d') \in P} (d'-b')}$ 
throughout. Note that this weighting function depends on the diagram, so that it is normalized in a weak sense. We use $\Sigma = 100I$ for the sublevel set featurization and $\Sigma = 25 I$ for the distance transform featurization (described in the next section), since the latter has a smaller maximum value. As with the Betti curves, we concatenate $0$ and $1$-dimensional curves to obtain a vector $\mathbf{v}_{\mathrm{Gaussian}} \in \mathbb{R}^{2N}$.

\subsection{Random Forests}
Random forests are built using decision trees. Decision trees assign outcomes (in our case, depth values) to inputs using axis-aligned decision boundaries, where the quality of a particular set of decision boundaries is evaluated through a loss function. Decision trees can be used to predict a scalar outcome using linear regression as the final stage, or they can be used to classify into predefined categories. We will use both versions in our experiments. Decision trees are non-parametric so they have low bias but high variability. A standard solution to try to minimize the total prediction error is to aggregate the results using a set of distinct decision trees, which together are called a \emph{random forest}. To create the random forest, new decision trees are created using bootstrapped samples of the data and a random subsample of features at each node, rather than using the entire dataset at every possible stage. We use random forests as our final stage in our pipeline since they require very little parameter tuning and also are capable of producing explainable results. 

\section{Overfitting and Generalization Challenges}\label{sec:cnn}

\begin{figure}[h]
    \centering
    \includegraphics[width=\linewidth]{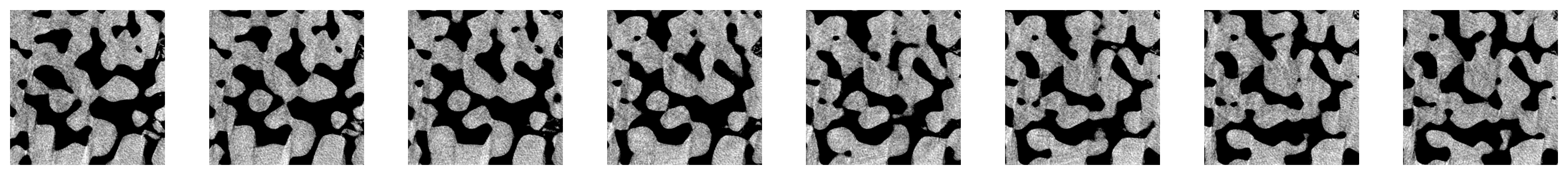}
    \caption{The first 8 samples from the 23m depth, showing a high degree of correlation in the overall shape of the ice and pore space.}
    \label{fig:correlationexample}
\end{figure}

Since the firn images do not have a fixed orientation, and are simply slices of a larger pattern, we want to ensure that any classifier trained on the data has outputs which are invariant to a large number of transformations, including rotation or translation of the image's field of view. An obstacle to this is that the data consists of slices that are taken close to one another, resulting in a high degree of correlation and low diversity in the images; see Figure~\ref{fig:correlationexample}. As a result, we expect a naive classifier to overfit to specific location-dependent features in the image.  

To confirm this hypothesis, we design a simple convolutional neural network (CNN), modeled on the Tensorflow image classification tutorial~\footnote[1]{\url{https://www.tensorflow.org/tutorials/images/cnn}}, and implemented using Tensorflow and Keras. Note that we are not attempting to establish the maximum performance of CNNs on this dataset; instead we are using a simple architecture to demonstrate how a conventional image classification approach can drastically overfit to the specific orientation of the images. The CNN archicture and details are shown in Figure~\ref{fig:CNN}. We make sure to incorporate at least two standard methods designed to prevent overfitting: dropout layers, and early stopping. 

After training on a training subset (75\% of the data), we test our CNN on three test sets: the remaining 25\% as-is, the remaining 25\% with a $3 \times 3$ Gaussian blur, and the remaining 25\% with a (potentially different) random symmetry of the square applied to each image. Results over ten trials are shown in Table~\ref{tab:cnnresults}. While the test accuracy on the original and blurred test set remains almost 100\%, the accuracy drops to 32\% for a test set with random symmetries applied. This suggests that the trained CNN classifier is far from invariant under reasonable transformations of the data. Various approaches could be taken to mitigate this defect, such as augmenting the training dataset with rotated and reflected images. However, this merely promotes a specific kind of invariance under a finite pre-defined set of transformations, and also significantly increases the amount of training data. Since our goal is a more robust approach which introduces far broader invariance properties, we turn to topological features in the next section.

\begin{figure}
\begin{tikzpicture}[ box/.style={ draw, rectangle, minimum width=2.5cm, minimum height=0.8cm }]
\node[box] (A1) {Input ($500 \times 500$)};
\node[box, below=of A1] (A2) {Convolution (32 filters, $3 \times 3$ kernel)}; \node[box, below=of A2] (A3) {Max Pooling ($2 \times 2$)};
\node[box, below=of A3] (A4) {Dropout (0.25)}; \node[box, below=of A4] (A5) {Convolution (64 filters, $3 \times 3$ kernel)};
\node[box, below=of A5] (A6) {Max Pooling ($2 \times 2$)}; 
\node[box, right=4cm of A1] (B1) {Dropout (0.25)};
\node[box, below=of B1] (B2) {Convolution (64 filters, $3 \times 3$ kernel)};
\node[box, below=of B2] (B3) {Fully Conected (64 units)};
\node[box, below=of B3] (B4) {Dropout (0.25)}; \node[box, below=of B4] (B5) {Fully Connected (10 units)}; 
\node[box, below=of B5] (B6) {Output (depth category)}; 
\foreach \x/\y in {A1/A2,A2/A3,A3/A4,A4/A5,A5/A6} 
\draw[->] (\x) -- (\y); 
\foreach \x/\y in {B1/B2,B2/B3,B3/B4,B4/B5,B5/B6} \draw[->] (\x) -- (\y); 
\draw[->] (A6.east) -| ($(A6.east)+(1.5,2.5)$) |- (B1.west); 
\end{tikzpicture}
\caption{Simple CNN architecture to evaluate overfitting on the firn dataset when no topological features are used. ReLU was used as the activation function for the hidden layers, and softmax for the output layer. We trained for a maximum of 50 epochs with batch size 20 and early stopping using a 20\% validation split, Adam as the optimizer, and with cross-entropy as our loss function.}\label{fig:CNN}
\end{figure}

\begin{table}[h]
    \centering
    \begin{tabular}{|l|c|}
    \hline 
    \multicolumn{2}{|c|}{Category prediction (\% accuracy) } \\ \hline
       \textbf{Unaltered test set}  &  $99.13 \pm 0.20$ \\ \hline
              \textbf{Test set with Gaussian blur}  & $99.07 \pm 0.22$  \\ \hline 
              \textbf{Test set with random symmetries applied }  &  $31.68\pm 1.05$ \\ \hline
    \end{tabular}
    \caption{Results of the CNN on different test sets. We show the mean and standard error over ten trials.}
    \label{tab:cnnresults}
\end{table}

\section{Design of Topological Features}\label{sec:methods}

\begin{figure}
    \centering
    \begin{subfigure}{\textwidth}
    \centering
    \begin{tikzpicture}[scale=0.7]
        \node at (0,-1.25) {\tiny Image};
        \node at (7,-1.25) {\tiny Sequence};
        \node at (12,-1.25) {\tiny Betti Curves};
        \node at (0,0) {\includegraphics[height=1cm]{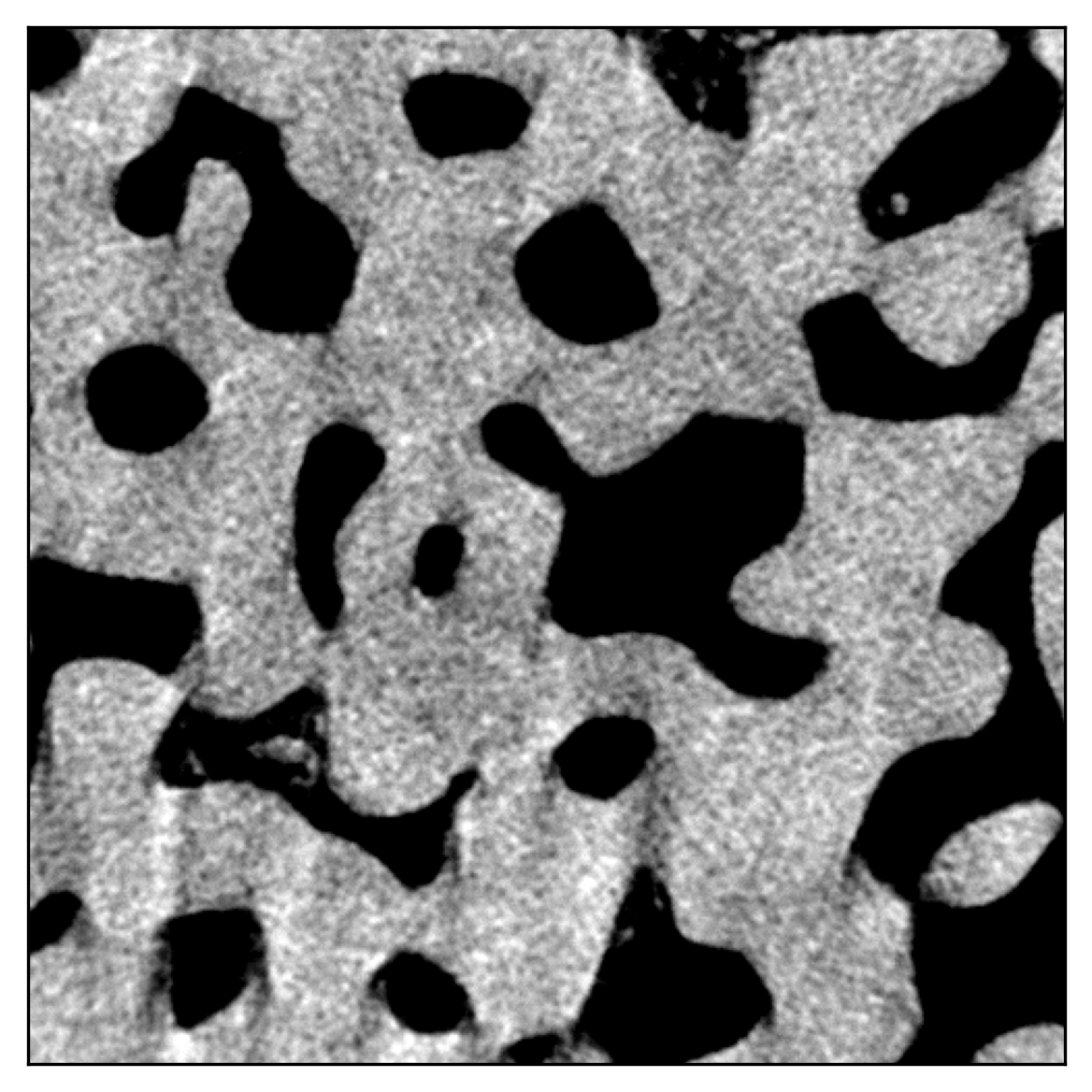}};
        \draw[->] (1,0)--(4.5,0);
        \node at (7,0) {\includegraphics[height=0.66cm]{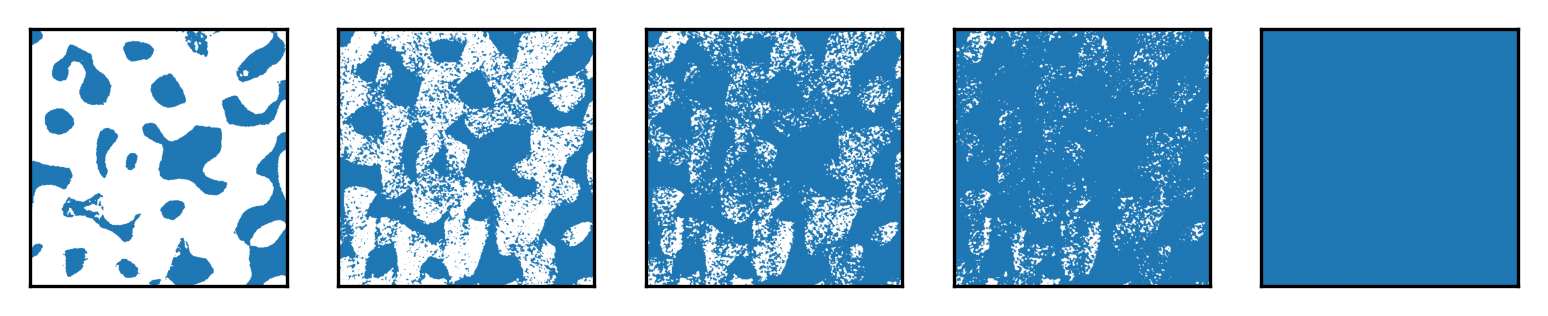}};
        \draw[->] (9.5,0)--(10,0);
        \node at (12,0.5) {\includegraphics[width=2.75cm]{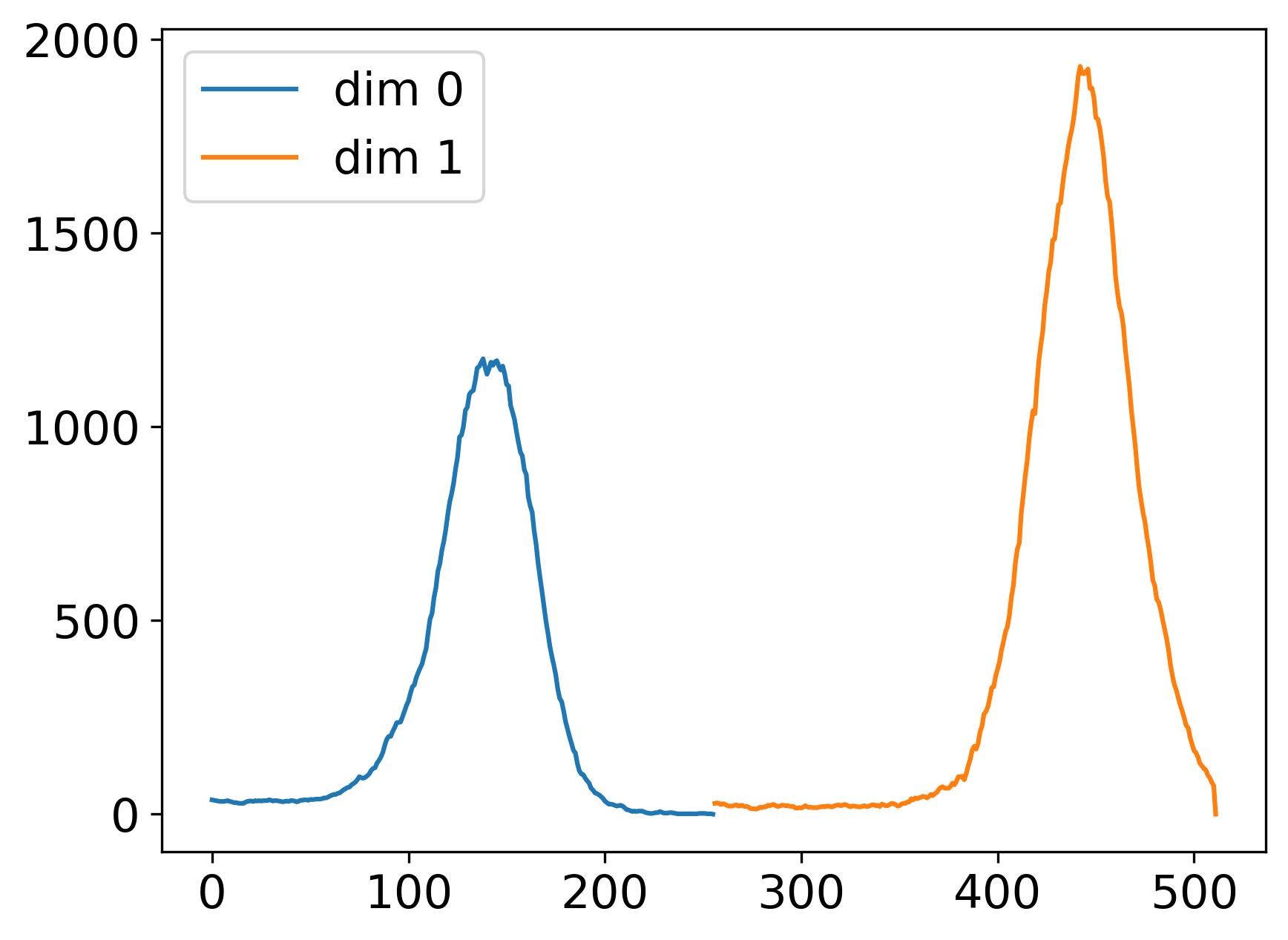}};
    \end{tikzpicture}
    \caption{Sublevel set featurization}
    \end{subfigure}

    \begin{subfigure}{\textwidth}
    \centering
    \begin{tikzpicture}[scale=0.7]
        \node at (0,-1.25) {\tiny Image};
        \node at (1.75, -1.25) {\tiny Binarized};
        \node at (3.5, -1.25) {\tiny D.T.};
        \node at (7,-1.25) {\tiny Sequence};
        \node at (12,-1.25) {\tiny Betti Curves};
        \node at (0,0) {\includegraphics[height=1cm]{images/pipeline/pipeline_original.png}};
        \node at (1.75,0) {\includegraphics[height=1cm]{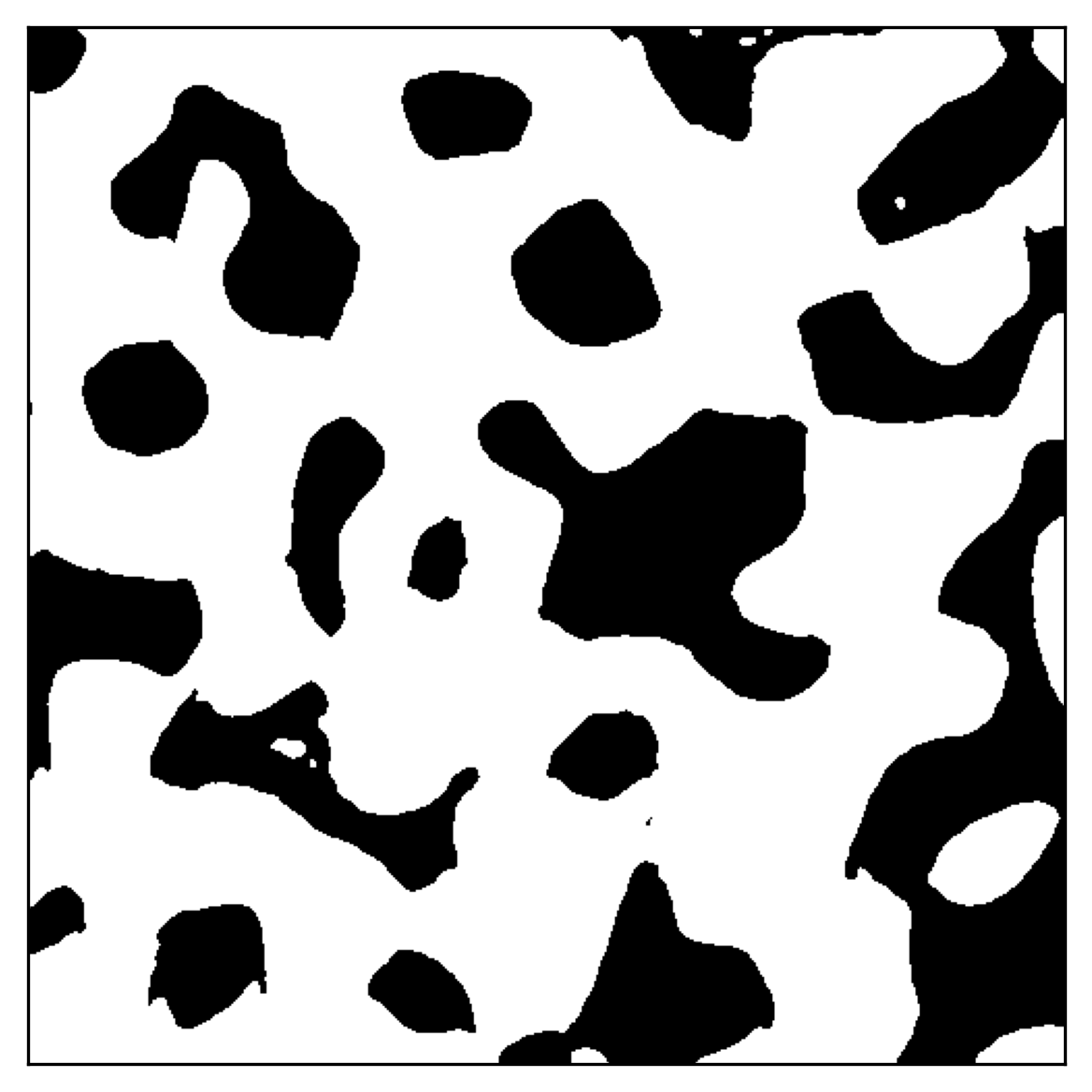}};
        \node at (3.5,0) {\includegraphics[height=1cm]{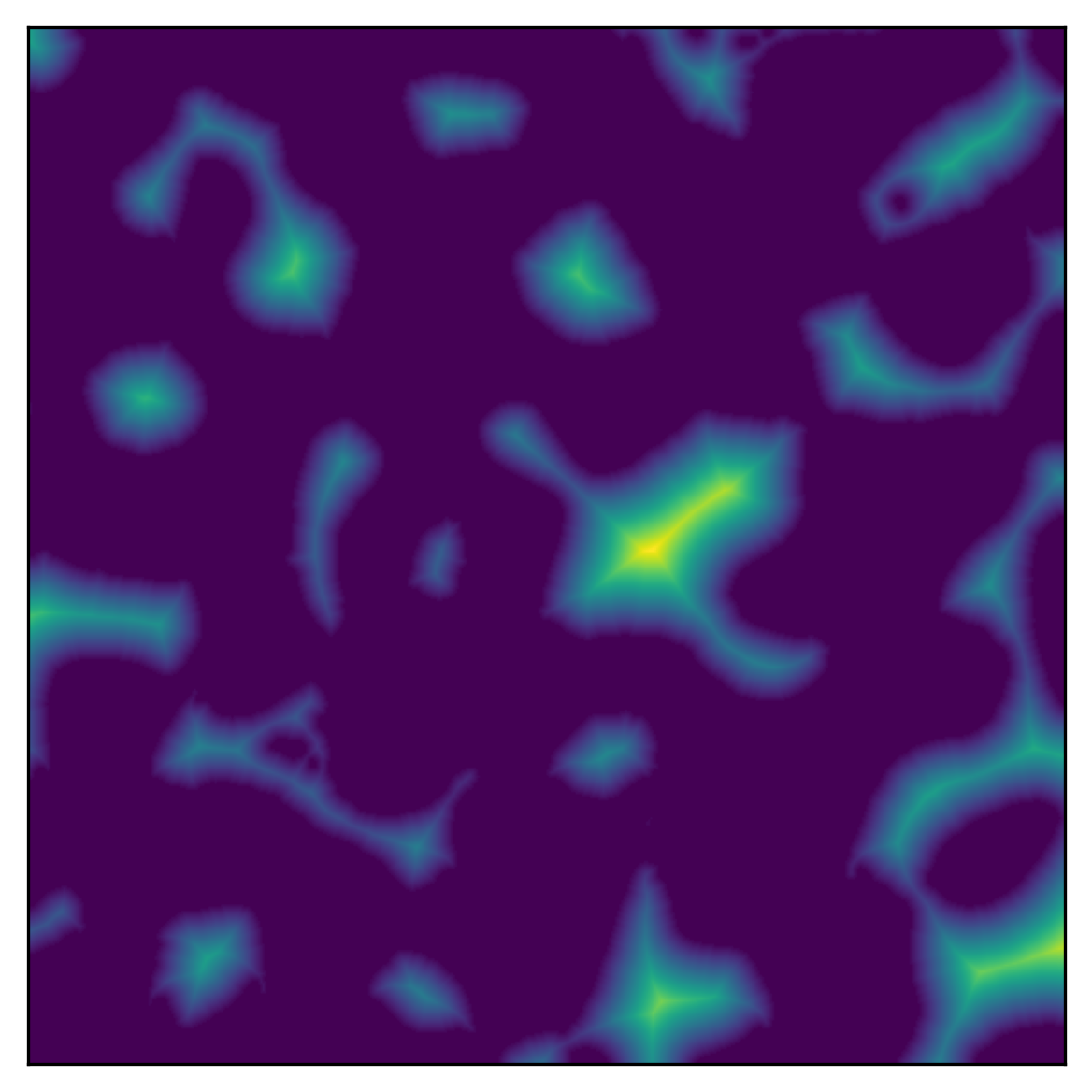}};
        \node at (7,0) {\includegraphics[height=0.66cm]{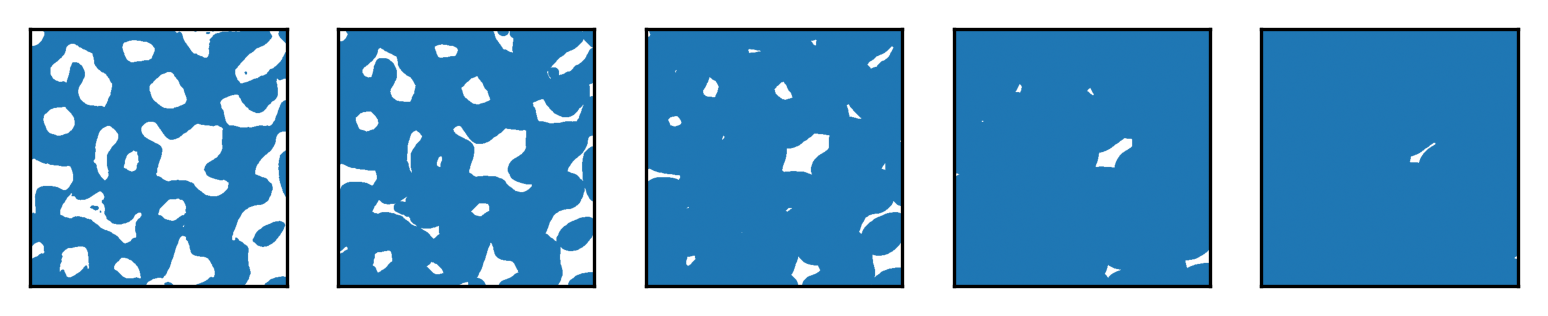}};
        \node at (12,0.5) {\includegraphics[width=2.75cm]{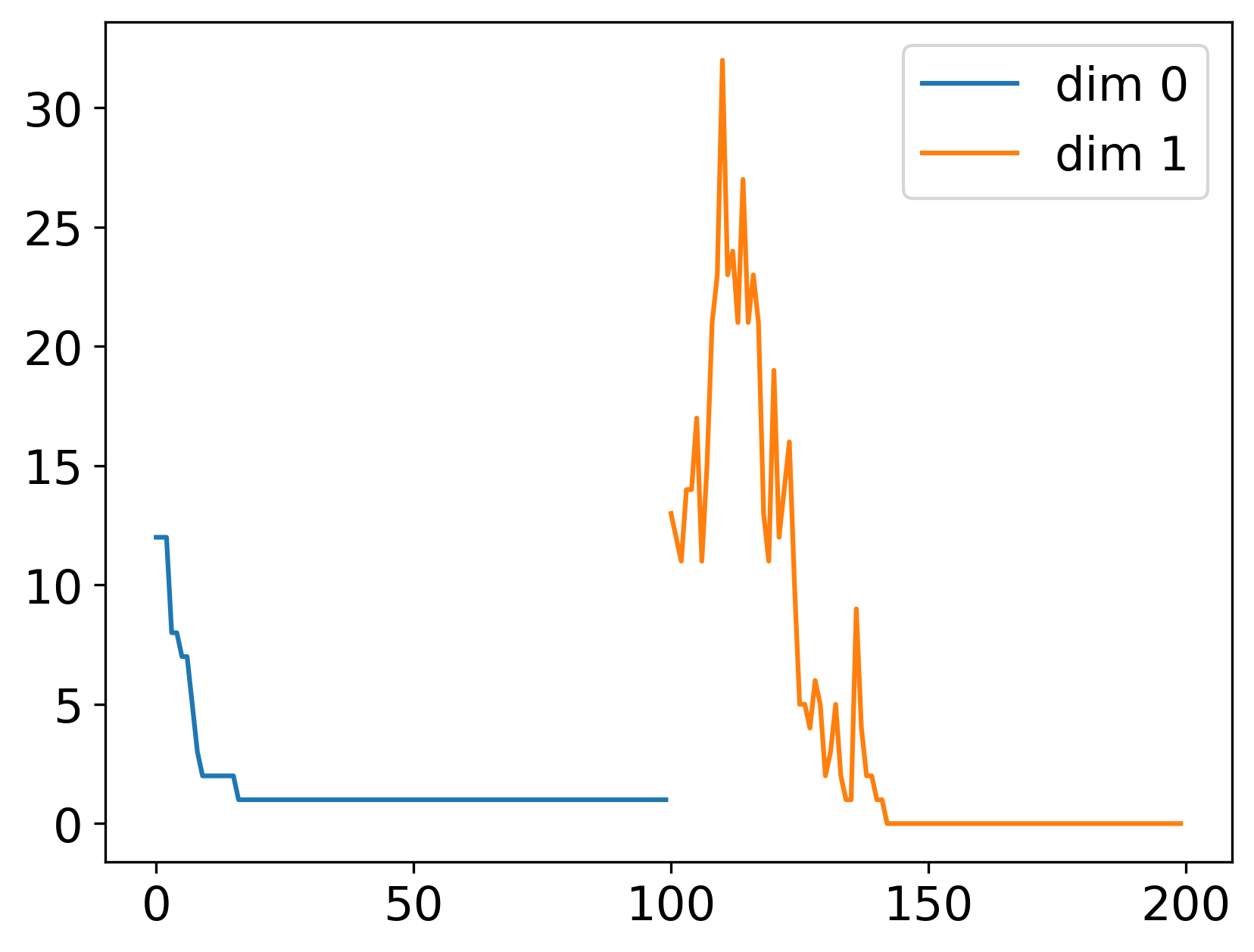}};
        \draw[->] (9.5,0)--(10,0);
        \draw[->] (0.75,0)--(1,0);
        \draw[->] (2.5,0)--(2.75,0);
        \draw[->] (4.25,0)--(4.5,0);
    \end{tikzpicture}
    \caption{Distance transform featurization}
    \end{subfigure}
    \caption{Visual representations of the two topological featurizations for an example image, using Betti curves as an example. In each sequence, blue indicates an included pixel in the cubical complex.}
    \label{fig:overview}
\end{figure}

In this section, we describe the two topological featurization methods used on the firn images as preprocessing steps, which we evaluate and compare in the next section using various experimental scenarios. In each case, persistent homology was computed using the \texttt{GUDHI} library's implementation of the cubical complex method in \cite{wagner2011efficient}. A visual overview of each method is given in Figure~\ref{fig:overview}.

\subsection{Sublevel Set Featurization}

Our first featurization method is inspired by the texture classification method in~\cite{chung2020smooth}. Each pixel in a grayscale image carries an intensity value from $0$ to $255$. Let $X_t$ denote the set of pixels with intensity at most $t$, so that $X_{255}$ is the entire image. Each image gives rise to a sequence of expanding pixel regions
$$
X_0 \subseteq X_1 \subseteq \cdots \subseteq X_{255}
$$
This data gives rise to two Betti curves (for $0^{th}$ and $1^{st}$ dimension) as described in Sections \ref{sec:homology} and Sections \ref{sec:curves} above, concatenated to give a Betti curve vector $\mathbf{v}_{\mathrm{Betti}}^{SS} \in \mathbb{R}^{512}$ and a Gaussian persistence curve $\mathbf{v}_{\mathrm{Gaussian}}^{SS} \in \mathbb{R}^{512}$. 

\subsection{Distance Transform Featurization}

Our second method is an implementation of the idea in \cite{hu2021two} to use a distance transform (DT) instead of raw pixel intensities. Each image is binarized using Gaussian smoothing and Otsu's binarization method~\cite{otsu1975threshold} so that pixels are assigned to either ice space or pore space. We define a function $f$ on the image as follows. On an ice space pixel, $f$ is defined to be zero. On a pore space pixel, $f$ is defined to be the Euclidean distance (in pixel space) to the nearest ice space pixel. Define $X_{DT}^t$ to be the set of pixels with $f$ value less than $t$:
$$
X_{DT}^t = \{ (i,j) \mid f((i,j)) \leq t \}
$$
This definition gives rise to a sequence of spaces 
$$
X_{t_1} \subseteq X_{t_2} \subseteq \cdots \subseteq X_{t_{N}}
$$
where $t_1 < t_2 <\cdot < t_N$ are the values in the range of $f$, rounded to integers. Note that this sequence is different to the sequence in the previous section. This sequence gives rise to a persistence curves, which we denote by $\mathbf{v}_{\mathrm{Betti}}^{DT}$ and $\mathbf{v}_{\mathrm{Gaussian}}^{DT}$ for Betti and Gaussian persistence curves respectively. In practice, we note that $f$ has a upper bound of $100$, so our resulting vectors lie in  $\mathbb{R}^{200}$.

\section{Classification Results for Topological Features}\label{sec:results}

\subsection{Experimental Setup}

Our input data consists of the images of firn at various depths, and the outcome to be predicted is the depth. As input features, we consider four options: the sublevel set featurization $\mathbf{v}^{SS}$ or the distance transform featurization $\mathbf{v}^{DT}$, with either Betti or Gaussian curves for each choice. On all featurizations, we use one of two types of predictor: a random forest regression method to predict depth as a scalar variable, or random forest classifier method to predict depth as a categorical variable with ten choices. We use \texttt{sklearn} for our implementation of these algorithms. Most of the default hyperparameters were utilized, only changing the split criterion in the classification task to entropy 

\begin{figure}[h]

\begin{subfigure}{0.3\textwidth}
{\includegraphics[width=\textwidth]{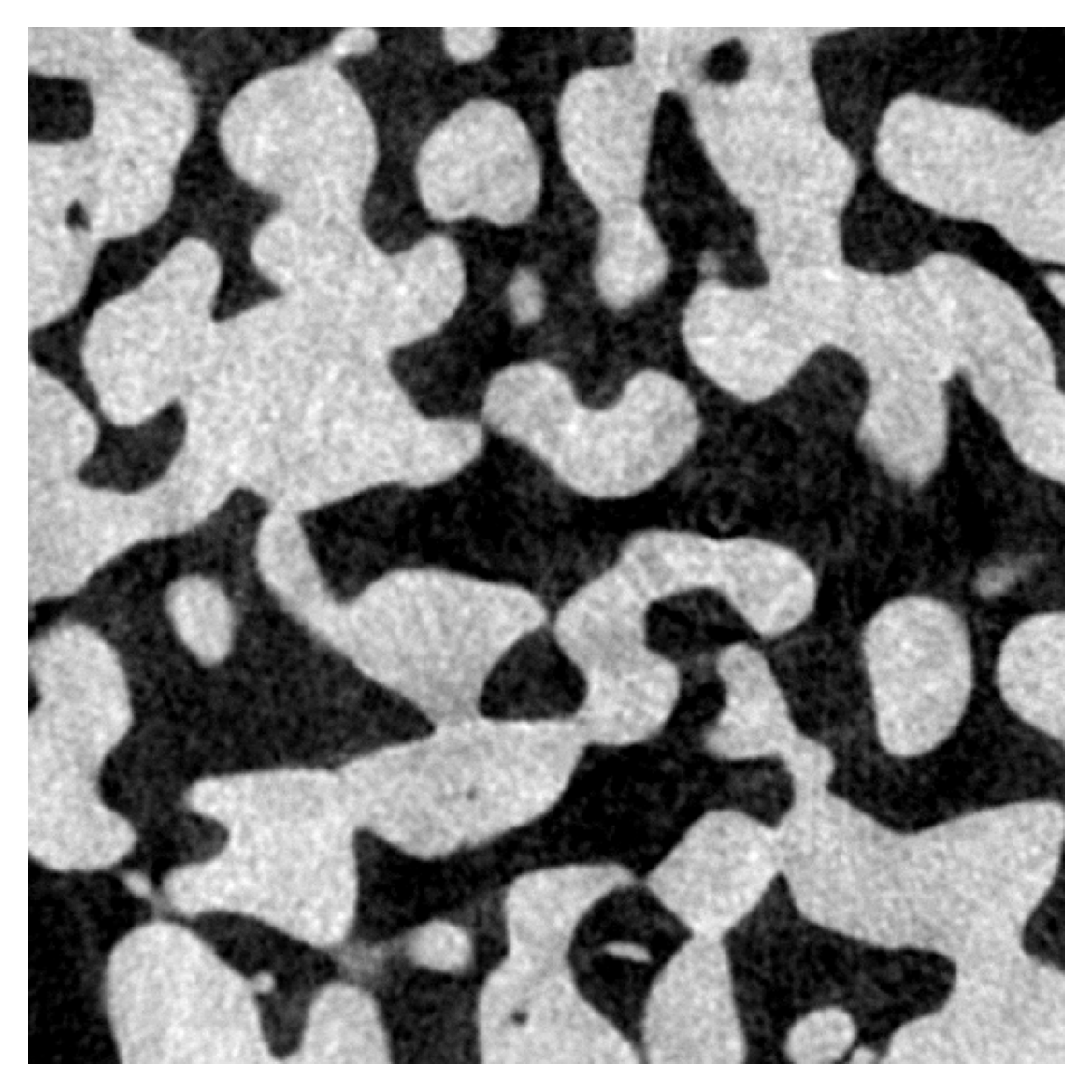}} 
\caption{Original}
\end{subfigure} \hfill
\begin{subfigure}{0.3\textwidth} 
{\includegraphics[width=\textwidth]{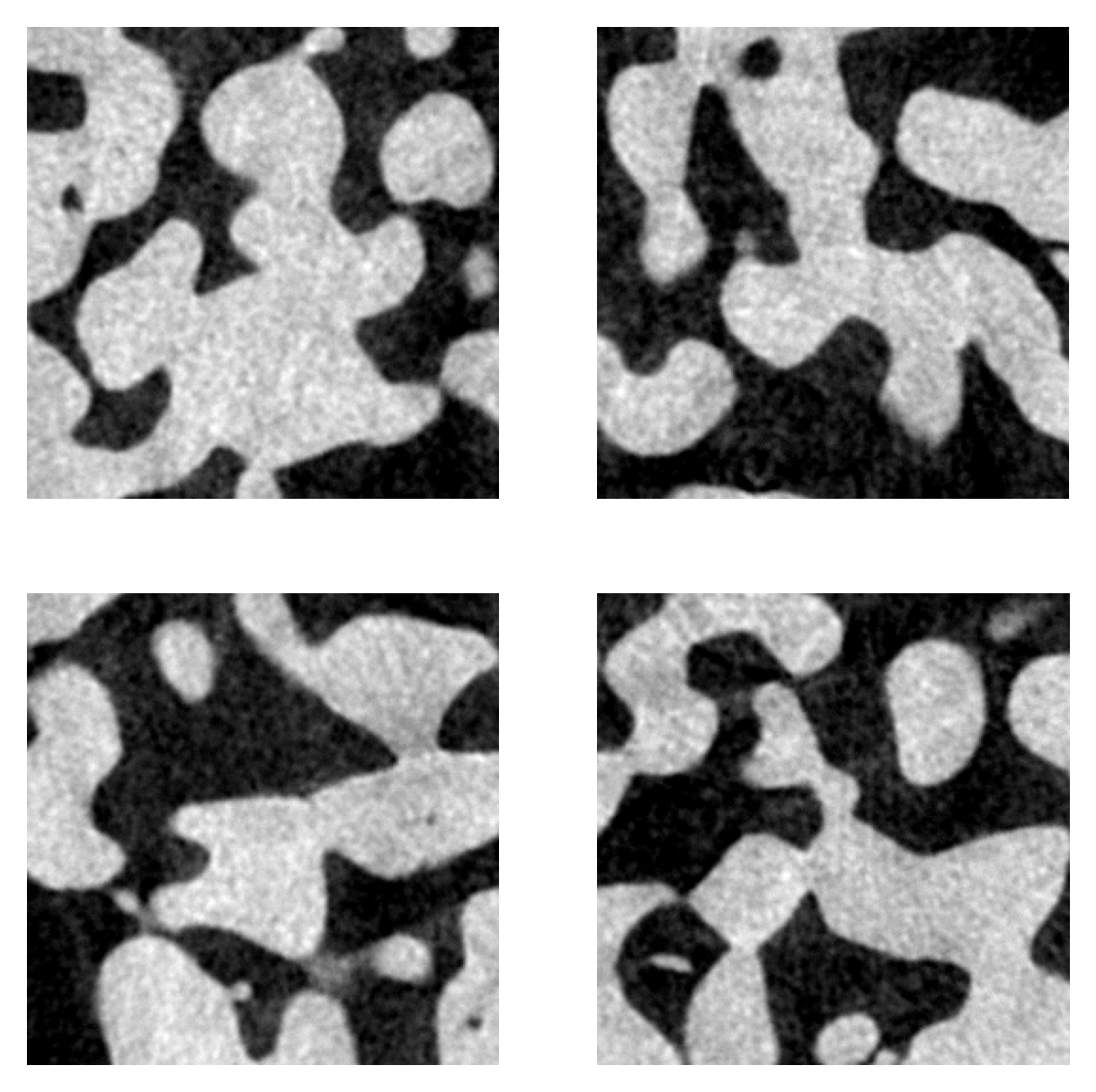}} 
\caption{Split}
\end{subfigure} \hfill
\begin{subfigure}{0.3\textwidth}
{\includegraphics[width=\textwidth]{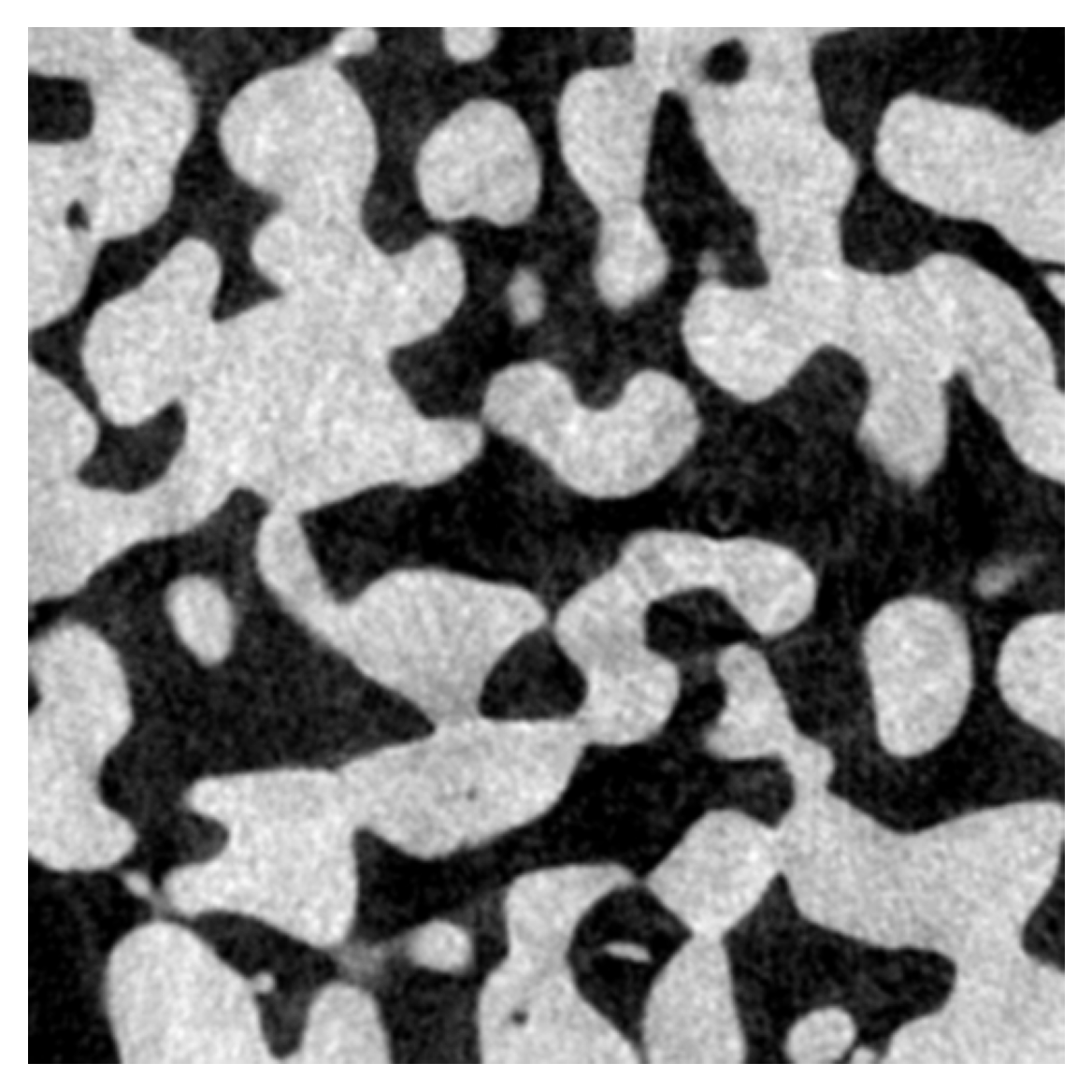}} 
\caption{Blurred}
\end{subfigure}
\caption{The original image and two manipulations used to evaluate predictor generalizability. The original image is shown in (a). In (b), we show the quadrants used for the \textbf{Split} and \textbf{Split BR} scenarios. The blurred version of the image (applied to the test set in \textbf{Blurred}) is shown in (c).}
\label{fig:Split}

\end{figure}

In order to test the generalizability of each setup, we create a range of challenging test-set scenarios:
\begin{itemize}
    \item \textbf{Whole:} We make a random $75\%/25\%$ train-test split of all images.
    \item \textbf{Split:} Each image is split into four quadrants. We then make a random $75\%/25\%$ test-train split of the set of all quadrant images. See Figure~\ref{fig:Split}.
    \item \textbf{Split BR:} Each image is split into four quadrants. Instead of a random train-test split, we assign all bottom right quadrants to the test set, and all others to the training set.
    \item \textbf{Blurred:} We make a random $75\%/25\%$ train-test split of all images. We then blur \emph{only} the test set with a $3\times 3$ Gaussian blur.
    \item \textbf{Missing depths:} The test set consists of full images from depths $23$m, $53$m and $70$m, while the training set consists of all other images. 
\end{itemize}

The \textbf{Split BR} and \textbf{Blurred} scenarios are designed to gauge the likelihood that the method would give reliable results on a different firn core in the future. The \textbf{Missing depths} scenario allows us to test the ability of the regression-based methods to predict out-of-sample depths.

\subsection{Results}

\begin{table}[h]
    \centering
    \begin{tabular}{|l|c|c|c|c|}
    \hline
         \multicolumn{5}{|c|}{Scalar prediction (mean absolute error in metres)}  \\
         \hline 
          & $\mathbf{v}_{\mathrm{Betti}}^{SS}$ & $\mathbf{v}_{\mathrm{Gaussian}}^{SS}$ & $\mathbf{v}_{\mathrm{Betti}}^{DT}$ & $\mathbf{v}_{\mathrm{Gaussian}}^{DT}$  \\
         \hline
      \textbf{Whole} &  0.74 $\pm$ 0.12  & 0.90 $\pm$ 0.14 & 4.01 $\pm$ 0.29 & 3.86 $\pm$ 0.25 \\
      \textbf{Split}  &  2.10 $\pm$ 0.10 & 1.92 $\pm$ 0.10 & 7.74 $\pm$ 0.15 & 8.36 $\pm$ 0.18 \\
        \textbf{Split BR} & 2.40 $\pm$ 0.04  & 2.21 $\pm$ 0.05 & 7.35 $\pm$ 0.02 & 7.86 $\pm$ 0.03   \\
      \textbf{Blurred} &  14.61 $\pm$ 0.78 &  11.26 $\pm$ 0.93 & 3.97 $\pm$ 0.27 & 3.97 $\pm$ 0.25 \\
      \textbf{Missing depths} &  9.93 $\pm$ 0.22 &  10.15 $\pm$ 0.26 & 6.20 $\pm$ 0.04 & 7.69 $\pm$ 0.10
      \\
      \hline
      \hline
         \multicolumn{5}{|c|}{Category prediction (\% accuracy)} \\
         \hline 
          & $\mathbf{v}_{\mathrm{Betti}}^{SS}$ & $\mathbf{v}_{\mathrm{Gaussian}}^{SS}$ & $\mathbf{v}_{\mathrm{Betti}}^{DT}$ & $\mathbf{v}_{\mathrm{Gaussian}}^{DT}$  \\
         \hline
      \textbf{Whole} & 99.03 $\pm$ 0.49 & 98.25 $\pm$ 0.63 & 71.72 $\pm$ 1.55  & 69.72 $\pm$ 2.24   \\
      \textbf{Split} & 95.95 $\pm$ 0.72 &  96.72 $\pm$ 0.27 &  45.98 $\pm$ 1.18 & 42.82 $\pm$ 0.93 \\
        \textbf{Split BR} & 92.02 $\pm$ 0.31  & 93.59 $\pm$ 0.21  & 46.22 $\pm$ 0.56  & 39.93 $\pm$ 0.43   \\
      \textbf{Blurred} &  50.59 $\pm$ 4.51 & 81.75 $\pm$ 1.61 & 71.34 $\pm$ 1.77 & 68.91 $\pm$ 2.04 \\
      \hline
    \end{tabular}
    \caption{Performance measures for topological feature-based depth prediction. When predicting depth as a scalar value we use mean absolute error (MAE). Note that true values range from $7$m to $78$m. For predicting depth as one of ten predefined levels (see Figure~\ref{fig:FirnDepths}), we use accuracy. Ten trials are performed to obtain each entry, with the mean and standard deviation shown.}
    \label{tab:mainresults}
\end{table}

Table \ref{tab:mainresults} shows the error (for scalar prediction) and accuracy (for category prediction) for the featurizations and train-test scenarios considered in this paper. For each setup, we run ten trials and report the mean and standard deviation. Overall, we see among the four featurizations, $\mathbf{v}_{\mathrm{Betti}}^{SS}$, $\mathbf{v}_{\mathrm{Gaussian}}^{SS}$, $\mathbf{v}_{\mathrm{Betti}}^{DT}$, and $\mathbf{v}_{\mathrm{Gaussian}}^{DT}$, none dominate in every category. Instead, the experiments reveal a number of interesting trade-offs. 

\subsection{Findings}\label{sec:findings}

We list some of the main findings here with brief explanations.

\subsubsection{Sublevel set features give very high accuracy on unblurred images.} This is likely due to the fact that the images in each depth category are highly correlated due to being slices taken close together. The sublevel set filtration method is able to leverage the small-scale information in each image to accurately predict the depth category.

\subsubsection{The performance of sublevel set features degrades on blurred test sets (with one exception).} 
\begin{figure}[h]
    \centering
    \begin{subfigure}{0.7\textwidth}
        \includegraphics[width=\textwidth]{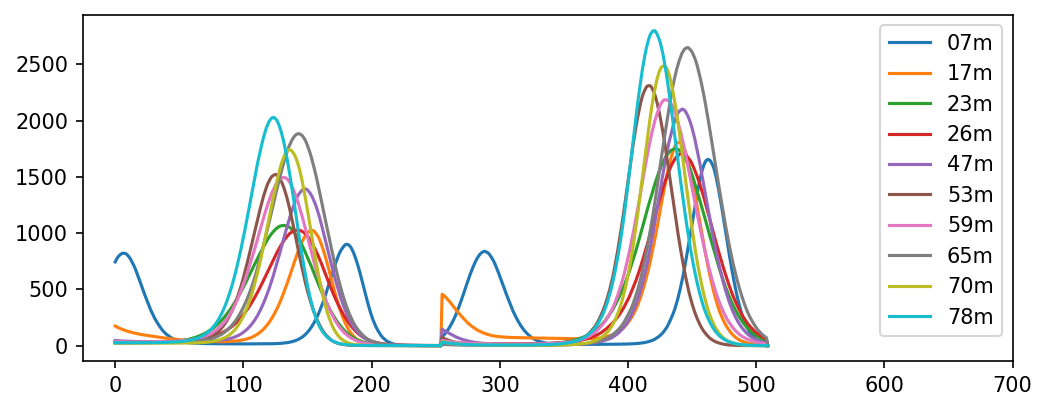}
        \caption{}
    \end{subfigure}
    \begin{subfigure}{0.2\textwidth}
        \includegraphics[width=\textwidth]{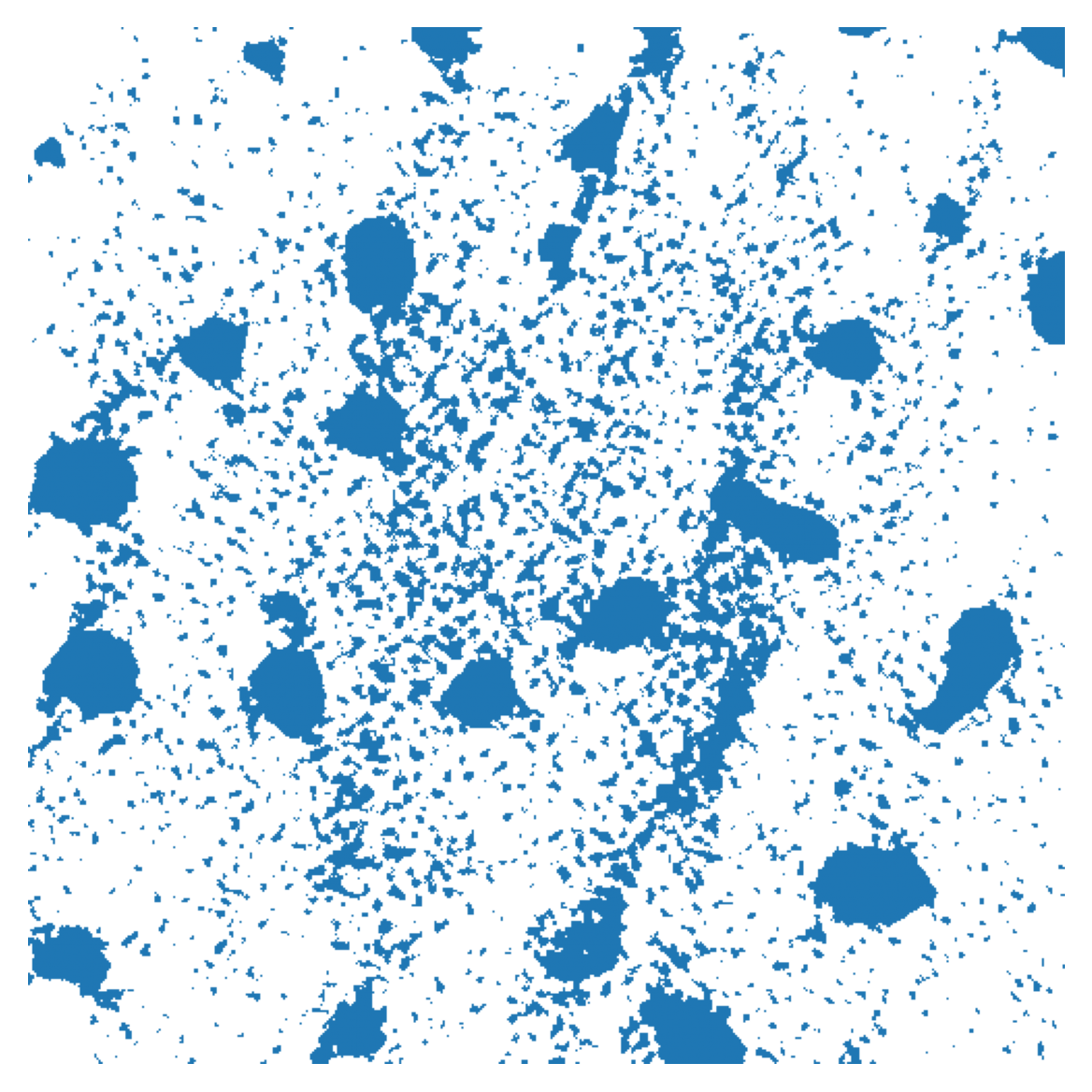}
        \caption{}
    \end{subfigure}
    \caption{On the left, the average Betti curve vector $\mathbf{v}_{\mathrm{Betti}}^{SS}$ by depth for the sublevel set featurization, showing good separation but conspicuously high peak values. On the right, we show a sample image from the $78$m depth threholded at $t = 120$, near the peak of the average curve for that depth. We see that the topology of the image at this threshold is dominated by granular noise.}
    \label{fig:allcurves}
\end{figure}

To understand why, see Figure \ref{fig:allcurves} which shows the mean $\mathbf{v}_{\mathrm{Betti}}^{SS}$ vector by depth. We see separation between depths, but observe that curves peak at values above 1,000. Recall that the Betti curve counts the number of holes or connected components at particular thresholds. Visually, we see that the pore space never has this many connected components in any sample, so the Betti curves are being dominated by smaller-scale structures. Figure \ref{fig:allcurves}(b), suggests that the Betti curve is detecting the granular noise in the ice space of the image. This will be confirmed in Section~\ref{sec:importance} below. This noise is altered by even small amounts of blurring, which explains the high error on blurred test images. The exception to this is $\mathbf{v}_{\mathrm{Gaussian}}^{SS}$ for categorical prediction. This is likely because the weighting function for the Gaussian persistence curve introduces a normalization factor which can account for the blurring. 

\subsubsection{Distance transform errors for scalar prediction are stable across tasks, and classification accuracy is robust to blurring.} For the distance transform features, we see that the mean absolute error is around $4$ for whole images, an error which corresponds roughly to the closest distance between adjacent depth categories. Blurring does not meaningfully affect the accuracy values for distance transforms, likely because the binarization is minimally affected.

\subsubsection{Distance transform performance is reduced on quadrant images.} This is likely because slicing up the image reduces the number of air pockets and increases the border effects, both of which can affect the distance transform and give less information to learn from. 

\subsubsection{Missing depths are better modeled by distance transform features than by sublevel set features.} Note that this is despite the sublevel set features being superior on the other unblurred test scenarios. This shows that the features detected by the distance transform features are easier to linearly extrapolate than the smaller-scale features detected by sublevel sets.

\section{Feature Importance and Explainability}\label{sec:featureimportance}

In addition to their theoretical invariance properties, another advantage of topological features is their explainability. In particular, Betti curves count connected components (for $\beta_0$) and holes (for $\beta_1$) at various thresholds, which in the case of firn has a physical meaning in terms of the ice-space and pore-space. Determining which topological features are most predictive can therefore be of independent interest since it can indicate which geological characteristics are most depth-dependent, and how these characteristics change with depth or other processes. By contrast, it is rarely possible to determine the exact signal being used by a convolutional neural network in a classification task. 

For each random forest classifier, we use the standard mean decrease in impurity (MDI) measure~\cite{breiman2001random} to rank features by importance. Recall that in our classifier, each feature corresponds to a choice of homological dimension ($H_0$ or $H_1$) and threshold. This threshold is a brightness threshold for the sublevel set featurization, or a distance threshold for the distance transform featurization. In either case, adjacent thresholds behave similarly, thus when choosing features to examine, we typically try to choose features whose thresholds are somewhat spaced out, as discussed below.

\subsection{Feature Importance for Sublevel Sets}\label{sec:importance}

We run a single trial of our random forest classifier with sublevel set features using whole, unaltered images (the \textbf{Whole} scenario from Table~\ref{tab:mainresults}). We then extract the top feature, and the next most important feature whose threshold value differs from that of the top feature by at least 25 (recall that brightness ranges from 0 to 255). This avoids us selecting two important features which operate on essentially the same brightness threshold.  The top feature was $\beta_0$ at threshold 128, and the next selected feature was $\beta_1$ at threshold 244. 

\begin{figure}[h]
    \centering
    \includegraphics[width=\linewidth]{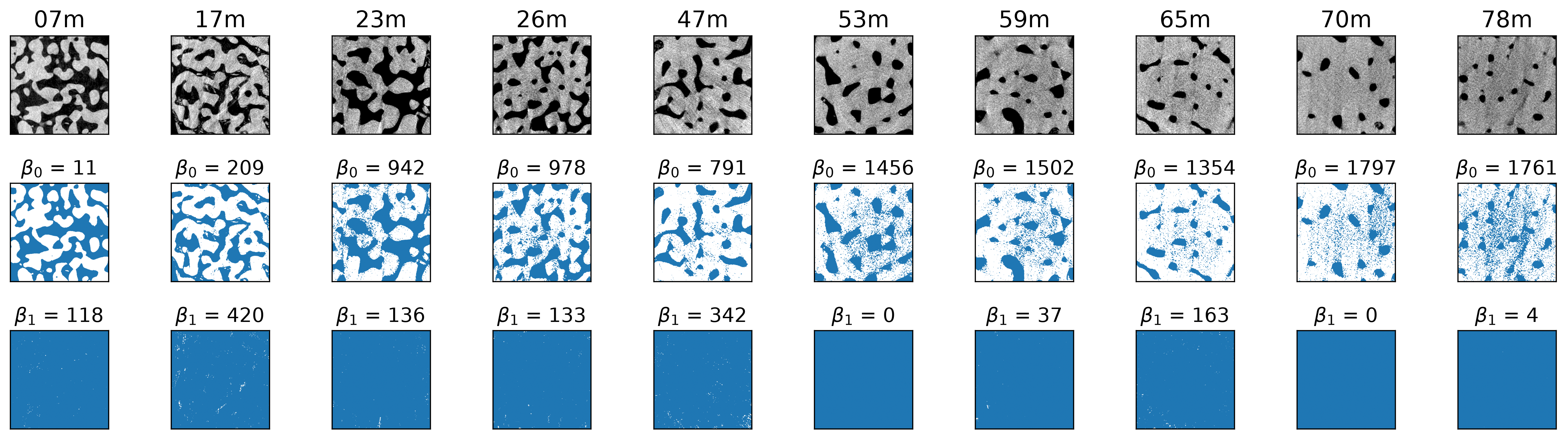}
    \includegraphics[width=0.9\linewidth]{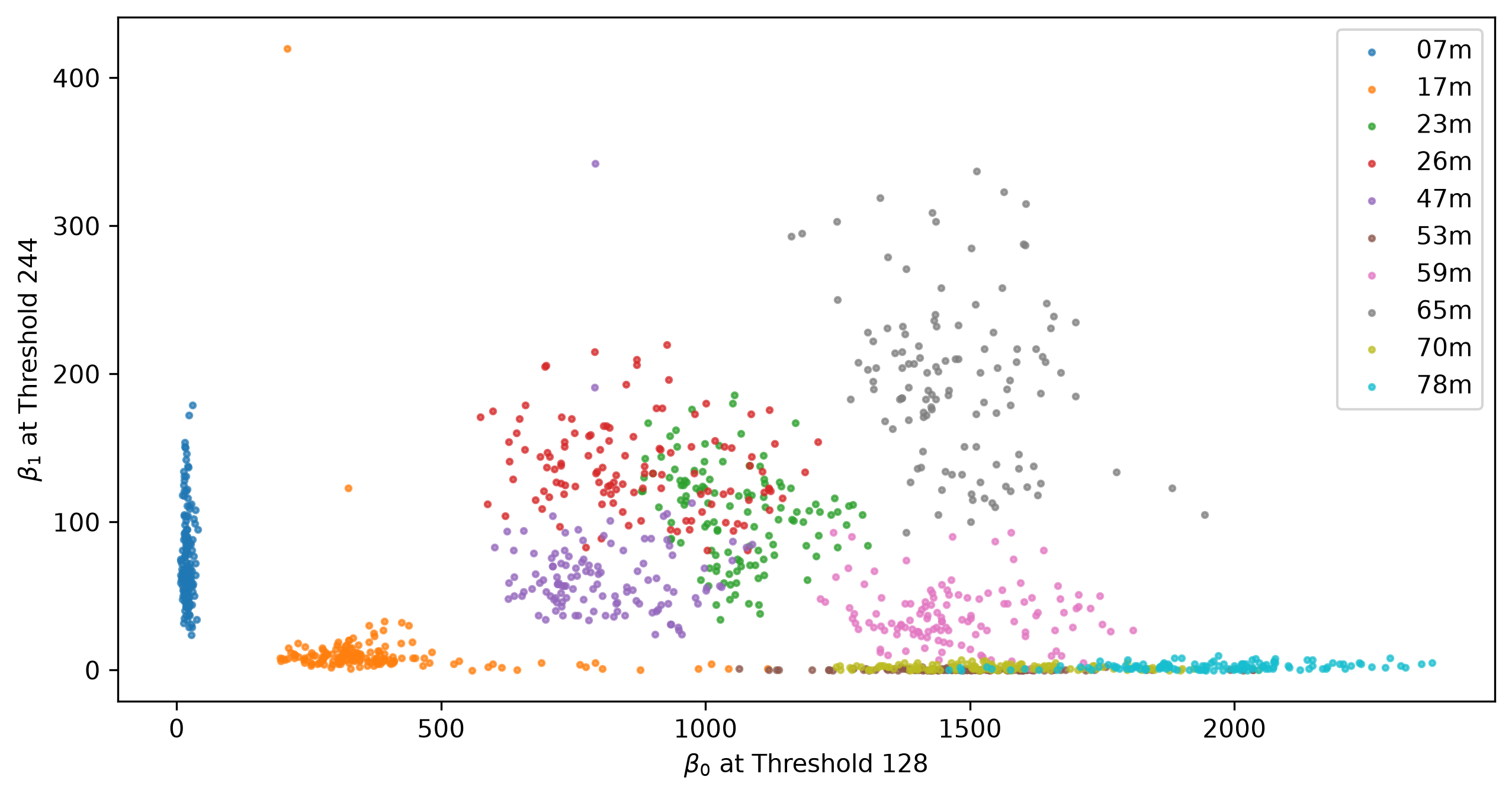}
    \caption{Examining two important features for the sublevel set featurization. We show sample images thresholded at the relevant thresholds with their corresponding Betti numbers (above). We also show a labeled scatter plot using the two features (below). }
    \label{fig:featureimportancesublevel}
\end{figure}

Figure~\ref{fig:featureimportancesublevel} shows the sample images from Figure~\ref{fig:FirnDepths} thresholded at the critical brightness thresholds (128 and 244), reporting their $\beta_0$ and $\beta_1$ numbers respectively. These results emphasize the finding from Section~\ref{sec:findings} and Figure~\ref{fig:allcurves} that the sublevel set featurization overfits to granular noise with many small features. Since this small-scale noise is not expected to be the same across different ice cores and microCT processes, this means that despite the good separation in Figure~\ref{fig:featureimportancesublevel}, this featurization is likely not generalizable, nor does it result in features which are interpretable in terms of geological characteristics.

\subsection{Feature Importance for Distance Transform}

As above, we run a single trial of our random forest classifier with sublevel set features using whole, unaltered images. We extract the top feature, and the next most important feature whose threshold value differs from that of the top feature by at least 10. Recall that for the distance transform, the threshold represents the distance (in pixels) from the ice-space, and our range was set to be $[0,100]$. The top feature was $\beta_0$ at threshold 0, and the next selected feature was $\beta_1$ at threshold 18.

It is of interest that the top feature was $\beta_0$ at threshold 0, since this is just the number of connected components of the ice space (gray/white area). Shallower depths tend to have a disconnected ice space, whereas deeper samples show a connected ice space surrounding small bubbles of pore space. Thus the random forest classifier confirms our intuition about what distinguishes shallow and deep samples. The next feature counts the number of holes when the ice space is expanded by 18 pixels (recall that the images are $500 \times 500$ pixels). This feature therefore measures the presence of deep pockets of air in the ice.

\begin{figure}
    \centering
    \includegraphics[width=\linewidth]{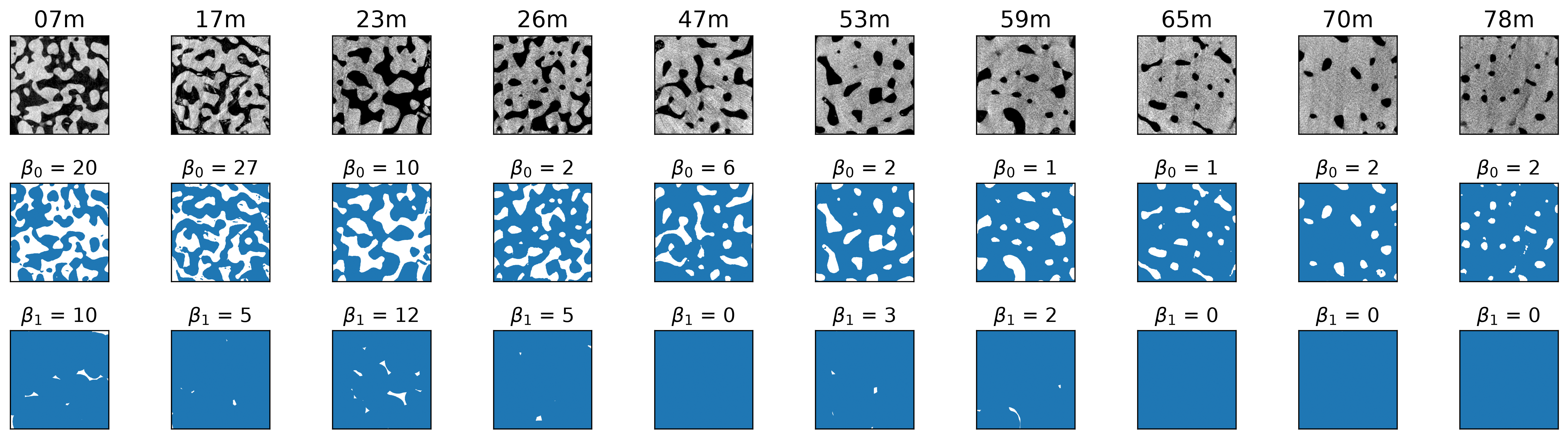}
    \includegraphics[width=0.9\linewidth]{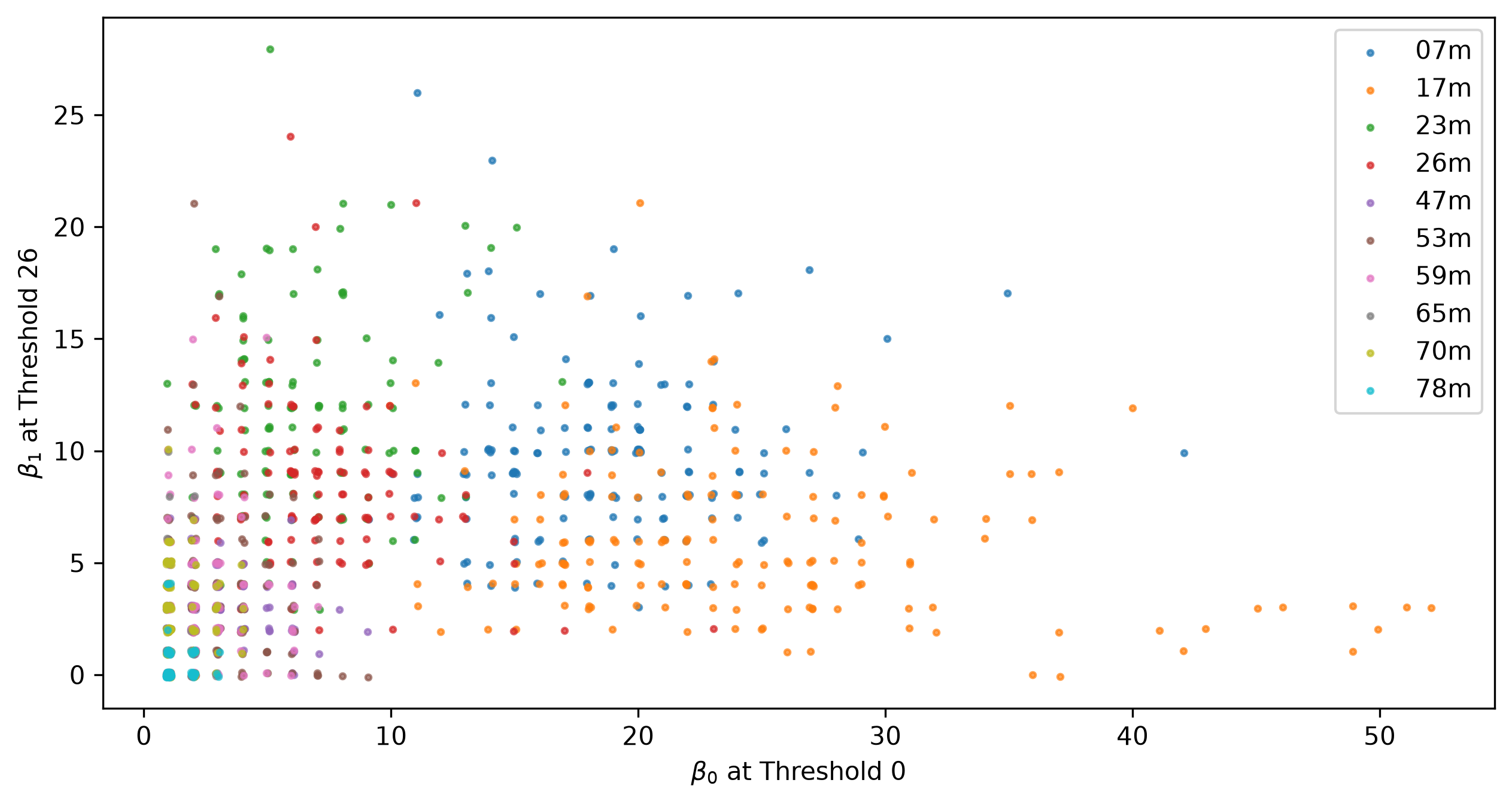}
    \vspace{0.1cm}
    \includegraphics[width=0.9\textwidth]{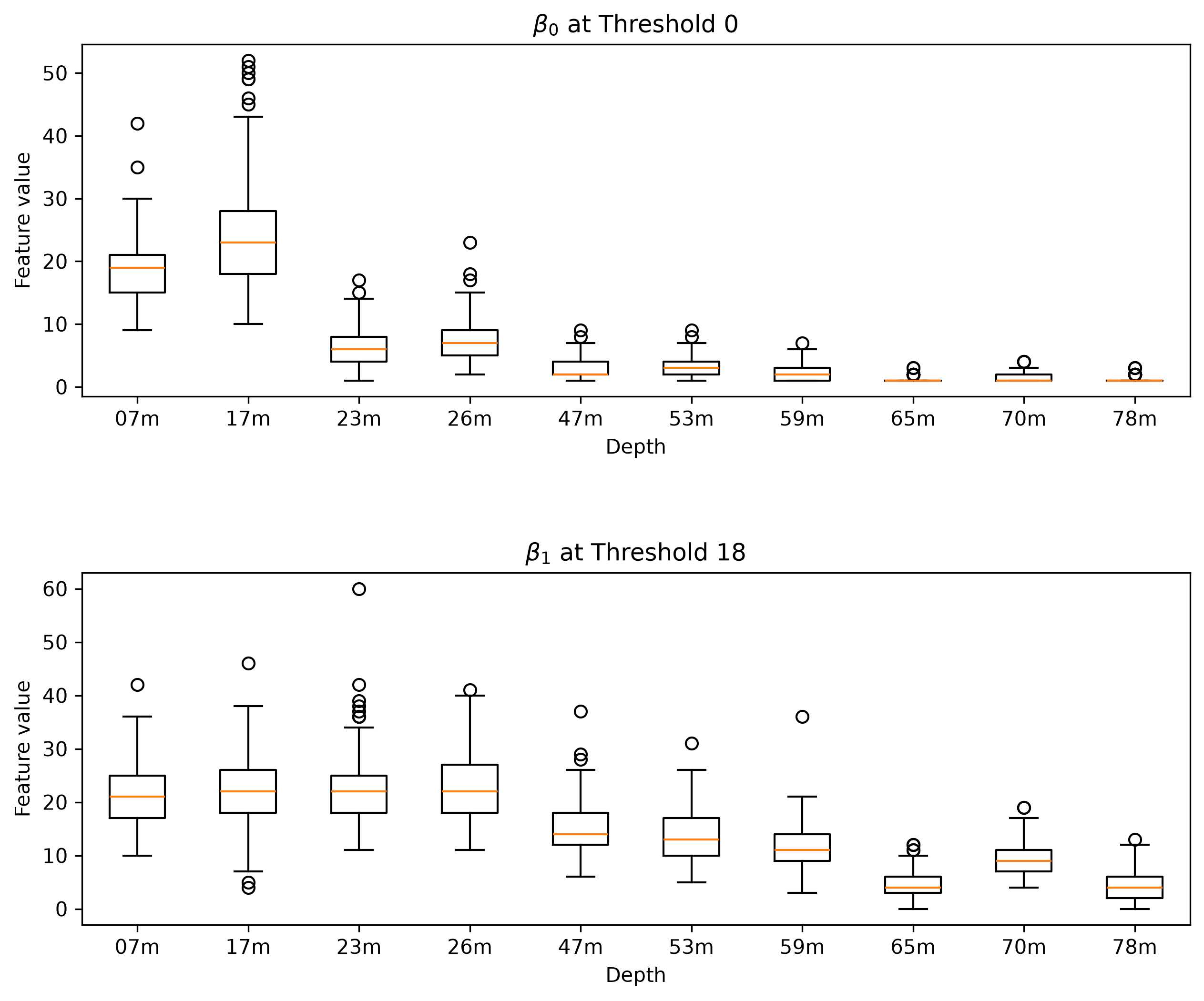}
    \caption{Examining two important features for the distance transform featurization. We show the distance transform for sample images thresholded at the relevant thresholds with their corresponding Betti numbers (top). We show a labeled scatter plot using the two features (middle), with tiny amounts of noise added so that multiple points at integer coordinates are visible. Boxplots of feature values by depth are also shown for each feature (bottom). }
    \label{fig:featureimportancedt}
\end{figure}

Figure~\ref{fig:featureimportancedt} shows sample images and a scatter plot for the distance transform features. We also provide boxplots to make the overall trends more visible. We see that the features are more interpretable in terms of the actual shape and size of the ice space in each sample image. From the boxplots, we see that $\beta_0$ at Threshold 0 (the connectedness of the ice space) separates the two bottom depths from the rest, while $\beta_1$ at Threshold 18 works to separate the depths into a 4-3-3 grouping.

\section{Conclusion}

Topological features have many advantages when attempting to analyze the kind of images in this paper, primarly that they are by design invariant to translation, rotation, and other transformations. In this paper, we sought to understand better what kinds of featurizations are better than others at various tasks. We find that in a straightforward classification task, sublevel set filtrations achieve almost perfect accuracy, but that their reliance on small scale features hinders their robustness to noise and their generalizability to out-of-sample data (i.e.~missing depths). This demonstrates that while the ``noise'' near the diagonal of persistence diagrams are often viewed as unimportant, a classifier based on topological features can be susceptible to overfitting to it. Merely using persistent homology to design features does not guarantee that the method will depend on the intended shape features. 

The distance transform, by contrast, allowed the classifier to fit to readily interpretable characteristics of the shape of the ice and pore space in each sample. This suggests that the classifier could generalize to different ice core datasets with images of similar size, but possibly different granular noise or illumination levels. The downside is that the distance transform captures much coarser information than the sublevel set filtration, and also depends on a separate binarization phase during preprocessing. This results in a controlled but significant error. However, this error is more stable to blurring effects and the distance transform classifier more readily predicts out-of-sample depths. 

We propose that future work on this challenging image processing task focus on features similar to those from the distance transform, reduce dependence on the binarization step (or use a more sophisticated binarization stage), and design more complex downstream classifier stages to improve accuracy while maintaining explainability. 

\section{Acknowledgements}

We would like to thank Xiaoli Gao for her crucial input and guidance on an earlier version of this project. 

\section{Conflicts of Interest}
The authors report no known conflicts of interest that could affect the outcomes of this study.

\bibliography{references}
\bibliographystyle{splncs04}

\end{document}